%% file: root.tex
\documentclass[letterpaper, 10 pt, journal, twoside]{IEEEtran}  

\IEEEoverridecommandlockouts                              

\usepackage{siunitx}
\usepackage[dvipsnames]{xcolor}
\definecolor{TS}{RGB}{230, 159, 0}
\definecolor{FS}{RGB}{53, 183, 121}
\definecolor{PS}{RGB}{49, 104, 142}
\definecolor{TO}{RGB}{68, 1, 84}

\usepackage{amsmath, amssymb, latexsym}

\usepackage{calc}

\usepackage{cite}

\usepackage{caption}
\usepackage{subcaption}
\usepackage{booktabs}

\usepackage[acronym]{glossaries}
\newacronym{rl}{RL}{Reinforcement Learning}
\newacronym{dr}{DR}{Domain Randomization}
\newacronym{da}{DA}{Domain Adaptation}
\newacronym{gan}{GAN}{Generative Adversarial Network}
\newacronym[firstplural={Markov Decision Processes (MDP)}]{mdp}{MDP}{Markov Decision Process}
\newacronym[firstplural={Partially Observable Markov Decision Processes (POMDP)}]{pomdp}{POMDP}{Partially Observable Markov Decision Process}
\newacronym{ppo}{PPO}{Proximal Policy Optimization}
\newacronym{psm}{PSM}{Patient Side Manipulator}
\newacronym{dvrk}{dVRK}{da Vinci Research Kit}
\newacronym{ui2i}{UI2I}{Unpaired Image-To-Image}
\newacronym{tr}{TR}{Tissue Retraction}
\newacronym{sofa}{SOFA}{Simulation Open Framework Architecture}

\usepackage{tikz}
\usetikzlibrary{backgrounds}
\usetikzlibrary{fit}
\usetikzlibrary{positioning} 
\usetikzlibrary{calc}
\usetikzlibrary{patterns} 
\usetikzlibrary{quotes} 
\usetikzlibrary{calc}
\usepackage{pgfplots}
\pgfplotsset{compat=1.16}
\usepgfplotslibrary{fillbetween}

\usetikzlibrary{fit}
\tikzset{
  fitting node/.style={
    inner sep=0pt,
    fill=none,
    draw=none,
    reset transform,
    fit={(\pgf@pathminx,\pgf@pathminy) (\pgf@pathmaxx,\pgf@pathmaxy)}
  },
  reset transform/.code={\pgftransformreset}
}

\usepackage{xspace}
\newcommand*{\eg}{\emph{e.g.}\@\xspace}

\newcommand*{\ie}{\emph{i.e.}\@\xspace}

\title{
Sim-To-Real Transfer for Visual Reinforcement Learning of Deformable Object Manipulation for Robot-Assisted Surgery
}

\author{Paul Maria Scheikl$^{1}$, Eleonora Tagliabue$^{2}$, Balázs Gyenes$^{1}$, Martin Wagner$^{3}$,\\ Diego Dall'Alba$^{2}$, Paolo Fiorini$^{2}$, and Franziska Mathis-Ullrich$^{1}$%
\thanks{Manuscript received: August 10, 2022; Revised November 16, 2022; Accepted November 27, 2022.}
\thanks{This paper was recommended for publication by Editor Pietro Valdastri upon evaluation of the Associate Editor and Reviewers' comments.
This work was supported by the Karlsruhe House of Young Scientists (KHYS), the Helmholtz Association under the joint research school ``HIDSS4Health – Helmholtz Information and Data Science School for Health", and the European Research Council (ERC) under the European Union's Horizon 2020 research and innovation programme under grant agreement No. 742671 (ARS).}
\thanks{$^{1}$ P. M. Scheikl, B. Gyenes, and F. Mathis-Ullrich are with the Institute for Anthropomatics and Robotics, Karlsruhe Institute of Technology, 76131 Karlsruhe, Germany. \newline {\small Corresponding author: \tt franziska.ullrich@kit.edu}}%
\thanks{$^{2}$ E. Tagliabue, D. Dall'Alba, and P. Fiorini are with the Department of Computer Science, University of Verona, 37134 Verona, Italy.}%
\thanks{$^{3}$ M. Wagner is with the Department for General, Visceral and Transplantation Surgery, Heidelberg University Hospital, 69120 Heidelberg, Germany.}%
\thanks{Digital Object Identifier (DOI): see top of this page.}
}

\begin{document}

\markboth{IEEE Robotics and Automation Letters. Preprint Version. Accepted November, 2022}
{Scheikl \MakeLowercase{\textit{et al.}}: Sim-To-Real for Visual RL in RAS} 

\maketitle

\begin{abstract}
Automation holds the potential to assist surgeons in robotic interventions, shifting their mental work load from visuomotor control to high level decision making.
Reinforcement learning has shown promising results in learning complex visuomotor policies, especially in simulation environments where many samples can be collected at low cost.
A core challenge is learning policies in simulation that can be deployed in the real world, thereby overcoming the sim-to-real gap.

In this work, we bridge the visual sim-to-real gap with an image-based reinforcement learning pipeline based on pixel-level domain adaptation and demonstrate its effectiveness on an image-based task in deformable object manipulation. We choose a tissue retraction task because of its importance in clinical reality of precise cancer surgery.
After training in simulation on domain-translated images, our policy requires no retraining to perform tissue retraction with a $50\%$ success rate on the real robotic system using raw RGB images.
Furthermore, our sim-to-real transfer method makes no assumptions on the task itself and requires no paired images.
This work introduces the first successful application of visual sim-to-real transfer for robotic manipulation of deformable objects in the surgical field, which represents a notable step towards the clinical translation of cognitive surgical robotics.
\end{abstract}


\section{INTRODUCTION}
    \label{sec:introduction}
    \IEEEPARstart{L}{earning} behaviors in simulation and transferring them to real robotic systems (sim-to-real) is a prominent topic of research in robot-assisted surgery since learning on a real surgical robotic system is often infeasible~\cite{thananjeyanMultilateralSurgicalPattern2017c, tagliabueSoftTissue2020, pore2021learning, shin2019autonomous}.
    Training in simulation enables end-to-end \gls{rl} of complex tasks in a safe and controlled environment, without requiring direct access to the real surgical robotic system.
    Training on real surgical robotic systems is impractical as \gls{rl} algorithms often require millions of environment interactions to train a policy and unsafe behavior during training may damage the system~\cite{pore2021learning}.
    
    \begin{figure}
        \centering
        \includegraphics[width=\columnwidth]{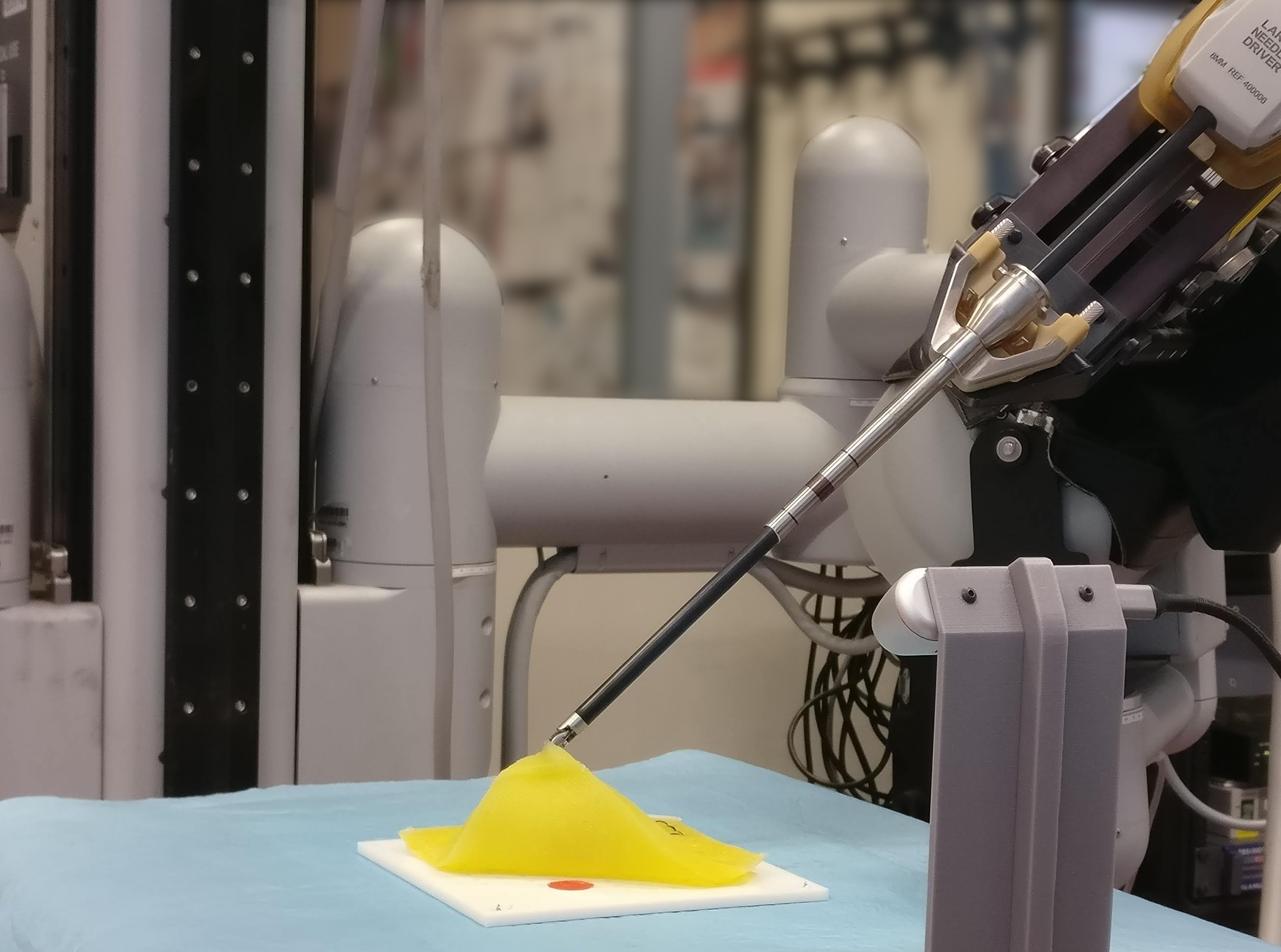}
        \caption{Experimental setup for tissue retraction. We combine an Intel RealSense camera and the da Vinci Research Kit with a ProGrasp instrument grasping a yellow sheet of silicone attached to a board.}
        \label{fig:overview}
    \end{figure}

    Existing works on sim-to-real policy transfer for robot-assisted surgery utilize low-dimensional observations such as the state of the robot and known goal positions, so that the inputs to the policy are the same during training in simulation and execution on the real robotic system~\cite{thananjeyanMultilateralSurgicalPattern2017c, tagliabueSoftTissue2020, dettorreLearningIntraoperativeOrgan2022}.
    The success of state-based policies heavily relies on accurate extraction of task-relevant information from the scene.
    This makes it highly challenging to exploit state-based methods in surgical applications involving the manipulation of deformable tissues.
    The large configuration space of deformable objects cannot be fully extracted from data provided by standard surgical sensors (\eg endoscopic camera) and is difficult to describe in a compact state.
    Image-based approaches, on the other hand, can learn task relevant features directly from sensor data.
    In this way, they are able to infer relevant information that is otherwise inaccessible to state-based approaches.
    However, image-based approaches have never been demonstrated in robot-assisted surgery due to the difficulty of transferring learned policies to real systems across the large visual domain gap between simulated and real images~\cite{scheiklCooperativeAssistanceRobotic2021, dettorreLearningIntraoperativeOrgan2022}.

    Several methodologies for transferring image-based policies across the visual domain gap between simulation and reality are currently investigated.
    \gls{dr} addresses the visual sim-to-real gap by randomly augmenting visual parameters of the simulation, \eg texture and lighting, such that the trained policy learns to extract generalized, task-relevant visual features.
    \gls{dr} approaches have been shown to translate well into reality~\cite{tobinDomainRandomizationTransferring2017, grunEvaluationDomainRandomization2019b, pengSimtorealTransferRobotic2018} but they can be highly task specific and hard to tune~\cite{openaiLearningDexterousInHand2019, matasSimtoRealReinforcementLearning2018}.
    In contrast, pixel-level \gls{da} addresses the visual sim-to-real gap by directly transforming the images from one domain into the other.
    Recent related works employ \glspl{gan} to transform images for robotic grasping tasks~\cite{hoRetinaganObjectawareApproach2021, raoRlcycleganReinforcementLearning2020, jamesSimtorealSimtosimDataefficient2019, bousmalisUsingSimulationDomain2018a}.
    These works share the limitation that large amounts of real world data are required to train the \glspl{gan}.
    Furthermore, application-specific auxiliary tasks must be defined in order to stabilize \gls{gan} training to avoid mode collapse and hallucinated objects in the translated images~\cite{raoRlcycleganReinforcementLearning2020}.
    This restricts their use to scenarios where these auxiliary tasks are applicable and may require additional data generation.
    Ho~et~al.~\cite{hoRetinaganObjectawareApproach2021} collect data from $135\,000$ task executions for \gls{gan} training.
    As an auxiliary task, they enforce predicting consistent bounding boxes by an object detection model for original and translated images.
    James~et~al.~\cite{jamesSimtorealSimtosimDataefficient2019} invert the problem by training a \gls{gan} to translate images from domain-randomized simulations into a canonical simulation and show that the \gls{gan} is also able to translate images from reality into the canonical simulation.
    Most of these works employ CycleGANs~\cite{zhuUnpairedImagetoimageTranslation2017} as their \gls{da} model.
    As an alternative to CycleGAN, CUT~\cite{parkContrastiveLearningUnpaired2020a} and DCL~\cite{hanDualContrastiveLearning2021} maximize mutual information between patches of the original and translated image through a contrastive learning approach.
    In robotics research, CUT is employed to generate synthetically labelled data for computer vision tasks such as semantic segmentation~\cite{imbusch2022synthetic, narasimhanSelfsupervisedTransparentLiquid2022} but has not been investigated for \gls{da} in \gls{rl}.\\

    In this work, we propose a pipeline for image-based \gls{rl} of a surgical robotic task.
    Sim-to-real transfer is achieved through pixel-level \gls{da} using a contrastive \gls{gan}.
    
    The contributions of this work are two-fold:

    \noindent
    1) In the field of surgical robotics, we present the first successful sim-to-real transfer of an image-based \gls{rl} policy to a real surgical robotic system.
        A visuomotor policy is trained in simulation and evaluated in reality without retraining.
        The approach is shown for \gls{tr} (see Fig.~\ref{fig:overview}), a common long-horizon task in deformable object manipulation.
        By relying on image inputs, the presented approach may be applied to other image-based tasks in robotic control, without the need of designing hand-crafted task-specific policies and is thus not limited to \gls{tr}.

    \noindent
    2) In the field of \gls{da}, we demonstrate that contrastive \glspl{gan} are capable of performing unpaired image-to-image translation with less data than previous methods and do not require additional application-specific auxiliary tasks to stabilize training.
        The contrastive loss serves as an auxiliary task to retain useful visual features during image translation.
        Unlike previous work, the contrastive loss is independent of the \gls{rl} task and can thus be considered a task-agnostic approach.
        This is the first successful application of pixel-level \gls{da} in \gls{rl} for deformable object manipulation.

\section{METHODS}
    \label{sec:methods}
    \input{plots/overview}
    The goal of this work is to train a visuomotor policy in a simulation of robot-assisted surgery on visual observations, such that the policy can be deployed on the real system without retraining.
    An overview of training and evaluation settings is given in Fig.~\ref{fig:training}.
    During policy learning, images from simulation $o_s$ are translated into the image domain of the real system $o_r$ before being passed to the policy $\pi$ that generates actions $a$ which are applied to the simulation.
    The required translation model $G$ is learned from unpaired examples of both the real and simulation domain and frozen during policy learning.
    Compared to related work, this presents a more general approach for sim-to-real transfer of visuomotor policies in deformable object manipulation as it does not require application-specific auxiliary tasks that stabilize training at the cost of restricting their transferability to other tasks.

    \subsection{Domain Adaptation}
        \label{sec:methods:da}
        A policy $\pi_s$ trained on simulated images is unsuitable for deployment in the real world because of the visual domain gap between observations in simulation and reality, even though the underlying dynamics and reward structure are similar.       
        Sequential decision making problems are frequently formalized as \glspl{mdp}.
        \glspl{mdp} can be generalized to \glspl{pomdp}~\cite{cassandra1994acting} that assume that the dynamics of the process are determined by an \gls{mdp}, while the true state of the process is hidden and cannot directly be observed.
        As the true state of the deformable object is not accessible, and we can only observe the scene through image observations, we formally interpret the \gls{tr} tasks in reality and simulation as two \glspl{pomdp} that differ in their transition $T$ (physical domain gap) and observation $O$ (visual domain gap) functions.
        Hidden states of the processes are assumed to be the same in simulation and reality.

        The goal of pixel-level \gls{da} is bridging the visual domain gap by changing the appearance of an image while task-relevant information is preserved.
        To this effect, this work investigates training an \gls{ui2i} translation model $G: O_s \rightarrow O_r$ that translates images $o_s$ from the simulation's observation function $O_s$, such that the translated images $\hat{o}_r = G(o_s)$ are indistinguishable from samples $o_r \sim O_r(o \, | \, \hat{z})$ taken from the real observation function $O_r$, where $\hat{z}$ is the hidden state.
        \gls{ui2i} methods utilize unpaired image data, making it feasible to train $G$ without access to hidden states of either environment.
        
        $G$ is trained on unpaired image data collected in simulation and on the real system such that different lighting conditions are represented in the dataset, and then frozen for policy training.
        A policy $\pi_g$ is then trained in simulation on translated observations $\hat{o}_r$.
        Thus, the policy $\pi_g$ can be directly deployed on raw images $o_r$ from the real camera during execution on the real robotic system.
        In order to encourage retaining features of real images, $G$ also learns an identity mapping of real images by minimizing the distance between real images $o_r$ and their translation $\hat{o}_r' = G(o_r)$.
        In contrast to previous work, this approach requires knowledge about the Cartesian positions of the robot's gripper, grasping point, and end point only during training in simulation and not during execution on the real system.
        
        Here, we investigate two methods for training contrastive \gls{ui2i} models, CUT~\cite{parkContrastiveLearningUnpaired2020a} and DCL~\cite{hanDualContrastiveLearning2021}, with CycleGAN~\cite{zhuUnpairedImagetoimageTranslation2017} as a baseline.
        Contrastive learning learns to associate a \textit{query} to a \textit{positive} example and to contrast the \textit{query} to \textit{negative} examples.
        In the case of CUT and DCL, the \textit{query} is a patch~\cite{parkContrastiveLearningUnpaired2020a} of the translated image, the \textit{positive} is the patch at the same location in the original image, and the \textit{negatives} are patches from the original image at other locations.
        Image fidelity is compared manually, with only the best method used for policy learning as translation model $G$.
        During policy learning, the instance of $G$ used at each time step is sampled uniformly from an ensemble of $7$ models from different training runs to compensate for possible bias of the individual \gls{gan} instances toward a specific lighting condition in the dataset.
        Intuitively, this is a form of visual domain randomization since the appearance of the translated images is slightly different for each specific instance of $G$ but the content of the images is the same.

    \subsection{Reinforcement Learning}
        \label{sec:methods:rl}
        \subsubsection{Tissue Retraction Task}
            \label{sec:methods:rl:tr}
            \begin{figure}[tb]
                \vspace{1.5mm}
                \centering
                \begin{subfigure}[b]{0.49\linewidth}
                    \centering
                    \begin{tikzpicture}[every node/.style={inner sep=0,outer sep=0}]
                    \node {\includegraphics[height=2.3cm]{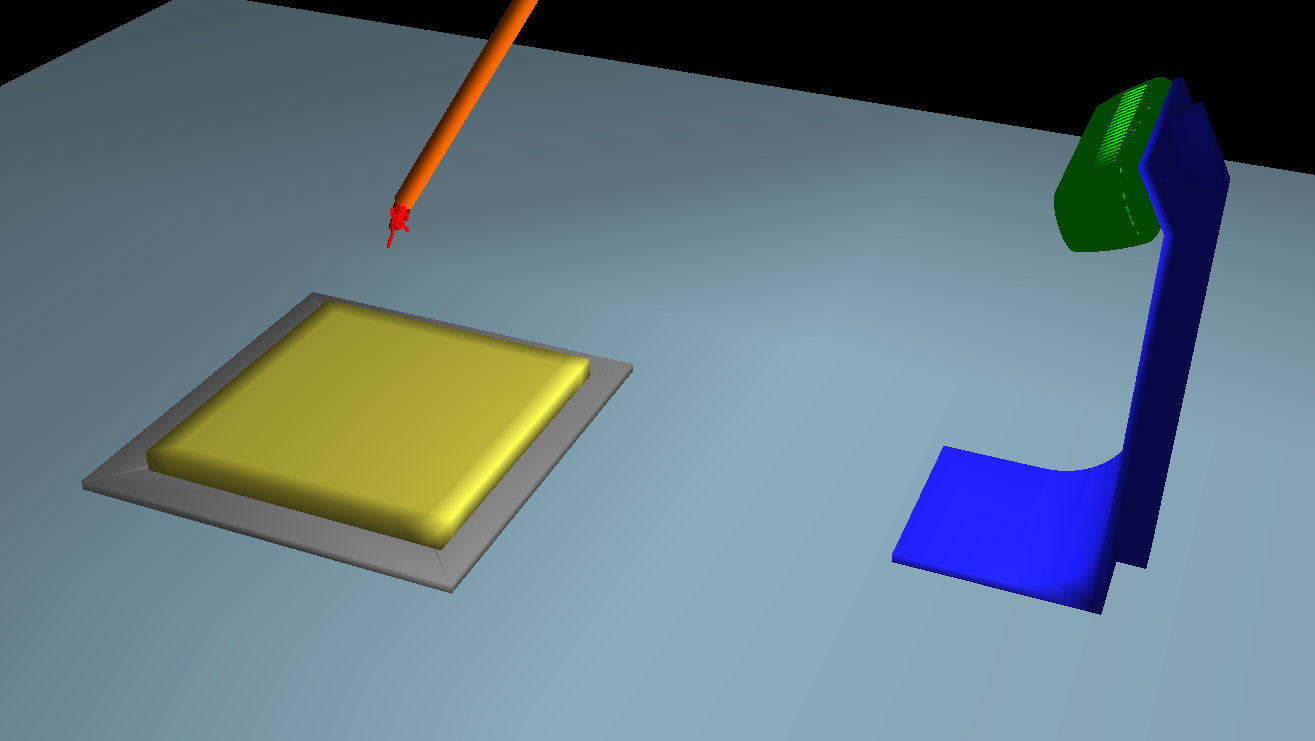}};
                    \draw[->, red, line width=0.8] (-0.75, -0.15) -- (-0.55, 0.05);
                    \draw[->, OliveGreen, line width=0.8] (-0.75, -0.15) -- (-0.45, -0.23);
                    \draw[->, blue, line width=0.8] (-0.75, -0.15) -- (-0.75, 0.15);
                    
                    \node[red, anchor=west] at (-0.55, -0.02) {x};
                    \node[OliveGreen, anchor=north] at (-0.65, -0.25) {y};
                    \node[blue, anchor=east] at (-0.85, 0.1) {z};
                    
                    \filldraw (-0.75,-0.15) circle (0.6pt);
                    
                    \end{tikzpicture}
                    \label{fig:sim}
                    \caption{}
                \end{subfigure}
                \hfill
                \begin{subfigure}[b]{0.49\linewidth}
                    \centering
                    \includegraphics[height=2.3cm]{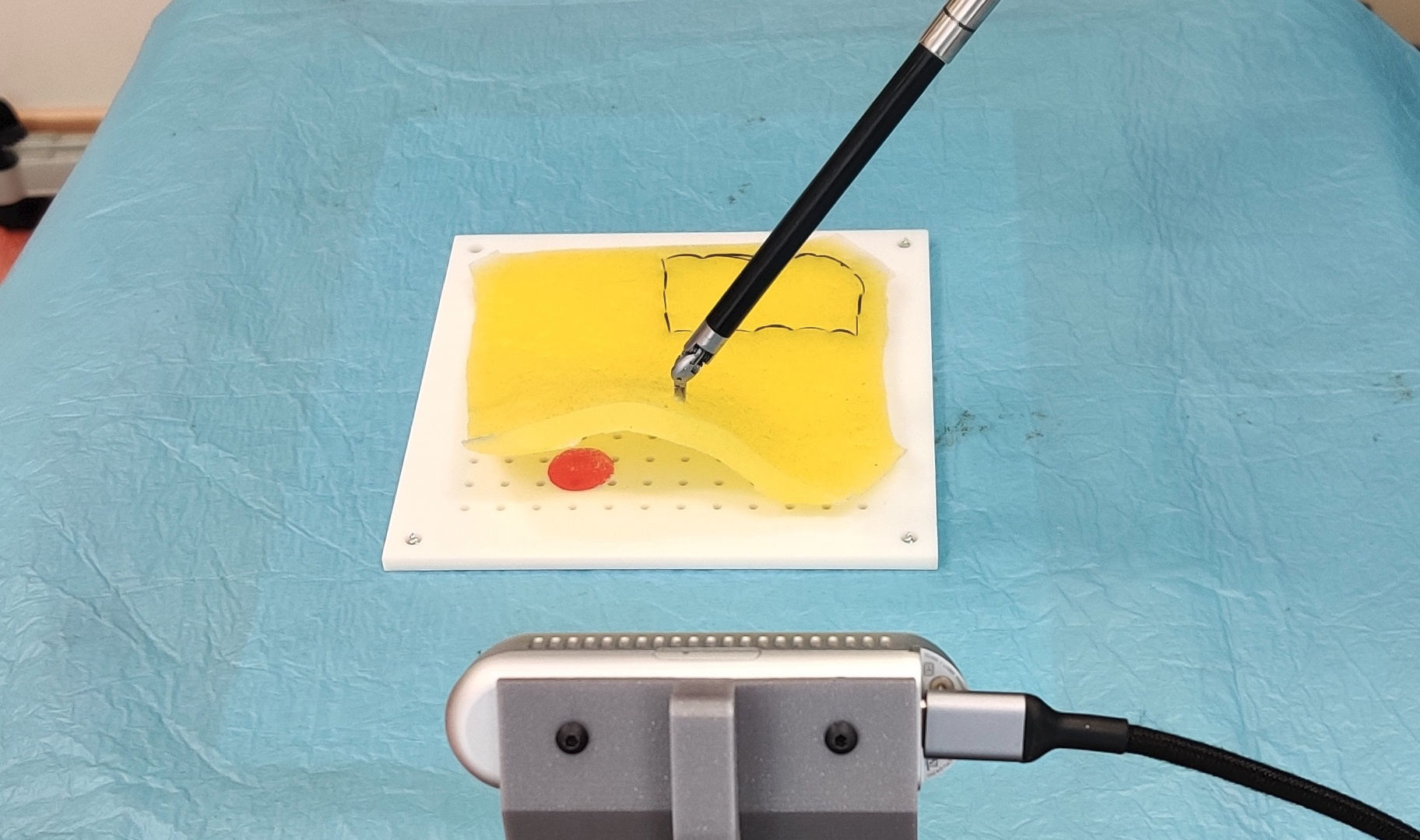}
                    \label{fig:real}
                    \caption{}
                \end{subfigure}
                \caption{(a) Tissue retraction scene implemented in SOFA with illustrated coordinate system and (b) experimental setup on the real robotic system.}
                \label{fig:task}
            \end{figure}
            \gls{tr} is an elementary surgical task that consists of grasping and pulling deformable tissue in order to expose a target area to the endoscopic camera.
            Furthermore, it is important to provide an appropriate force to tension the tissue for dissection without ripping tissue apart.
            \gls{tr} is thus of utmost importance, especially in cancer surgery, in order to dissect cancerous lymphatic tissue off of major blood vessels. 
            Due to its ubiquity in surgical procedures, autonomous execution of \gls{tr} has previously been investigated in the literature~\cite{tagliabueSoftTissue2020, pore2021learning, pore2021safe, attanasio2020autonomous}.
            In this work, we implement \gls{tr} of a rectangular soft tissue~\cite{tagliabue2022deliberation}.
            The position of the target, as well as the attachment points to the surrounding environment, are assumed to be known from pre-operative data to limit complexity, even if in reality they will change during surgical dissection. 
            The task is executed on the \gls{dvrk} using a single \gls{psm}, as illustrated in Fig.~\ref{fig:task}. 
        
        \subsubsection{Learning Environment}
            \label{sec:methods:rl:env}
            The \gls{tr} task is implemented in the \gls{sofa}~\cite{faureSOFAMultimodelFramework2012} framework (see Fig.~\ref{fig:task} (a)), relying on a finite element method.
            The tissue is modeled as a mesh of tetrahedral elements with its top right part fixed to simulate the attachment to the board in the real environment.
            The robot is modeled as the distal part of a \gls{dvrk} \gls{psm} including the gripper and instrument shaft respecting a fixed remote center of motion.
            Motion of the robot is constrained to a \SI{83}{\mm} high, \SI{180}{\mm} deep, and \SI{140}{\mm} wide workspace box above the tissue similar to the typical operative space of the \gls{psm}.
            The gripper starting position in each episode is uniformly sampled from the workspace with a minimum height of \SI{40}{\mm}.
            
            The \gls{tr} task is divided into grasping and retracting phases.
            The positions of the grasping and end points, as well as the target, are fixed for training and evaluation.
            The grasping point was selected above the target area, while the end point was chosen to achieve good visibility of the target area in the image observations, similar to previous work~\cite{tagliabueSoftTissue2020}.
            
            In the grasping phase, the gripper can move freely in the workspace.
            If the distance between the gripper and the desired grasping point decreases below \SI{3}{\mm} and the gripper is below the tissue surface, the tissue is automatically grasped and the retraction phase is entered.
            In the retraction phase, the tissue is attached to the gripper.
            The episode is completed successfully when the distance between the gripper and the desired end point reduces to \SI{3}{\mm} or less.
            Each episode is limited to $1\,000$ steps before the environment is automatically reset.
            Collisions are detected between the gripper jaw tips and the tissue, based on whether the gripper is within a bounding box around the tissue in its non-deformed state.
    
        \subsubsection{Reward, Observation and Action Space}
            \label{sec:methods:rl:spaces}
            The simulation environment adheres to the OpenAI gym standard~\cite{brockman2016openai}.
            The policy outputs the parameters of a Gaussian distribution, from which actions consisting of task space velocities are sampled.
            The continuous action is clipped to the interval~$[-1, 1]$, scaled such that the limits of the action space correspond to a maximum robot velocity of \SI{3}{\mm\per\s} in each direction, and an action repeat of $3$ is applied.
            The environment generates a $256\times256$ RGB image from a static camera perspective, which is then translated by the image-to-image translation model $G$.
            The four most recent images are concatenated and scaled down to a resolution of $84\times84$, resulting in observations of size $84\times84\times12$ that are subsequently passed to the policy.
            Observation space and network architecture are based on~\cite{mnih2015human}.
            
            The reward function is split into grasping $r_g$~(Eq. \ref{eq:rg}) and retraction $r_r$~(Eq.~\ref{eq:rr}) phases to match the two phases of the \gls{tr} task.
            In the grasping phase, the agent receives a negative reward proportional to the distance between the current gripper position $\boldsymbol{x}_t$ and the grasping point $\boldsymbol{x}_g$, normalized by the absolute size of the workspace:
            \begin{equation}
                d_g = - \dfrac{||\boldsymbol{x}_g - \boldsymbol{x}_t||_2}{||\boldsymbol{x}_{max} - \boldsymbol{x}_{min}||_2}
            \end{equation}
            where $\boldsymbol{x}_{max}$ and $\boldsymbol{x}_{min}$ are the corner points of the workspace bounding box.
            It further receives a constant negative reward equivalent to the reward seen at the beginning of the retraction phase based on grasping point $\boldsymbol{x}_g$ and end point $\boldsymbol{x}_e$:
            \begin{equation}
                c_g = - \alpha*\dfrac{||\boldsymbol{x}_g - \boldsymbol{x}_e||_2}{||\boldsymbol{x}_{max} - \boldsymbol{x}_{min}||_2}.  
            \end{equation}
            A weight of $\alpha = 1.2$ ensures that the agent is incentivized to transition to the retraction phase.
            In order to enforce safe policy behaviors on the real robot, an additional term $p_c$ penalizes collisions between gripper and tissue proportional to the distance to the grasping point, and a further term $p_w$ penalizes actions that would violate the workspace boundaries.
            Finally, the reward function assigns a one-time positive reward of $e_g = 1$ to a successful grasp.
            In the retraction phase, the agent receives a negative reward $d_e$ proportional to its distance to the end point, normalized by the absolute size of the workspace: 
            \begin{equation}
                d_e = -\dfrac{||\boldsymbol{x}_e - \boldsymbol{x}_t||_2}{||\boldsymbol{x}_{max} - \boldsymbol{x}_{min}||_2}.
            \end{equation}            
            Additionally, a one-time positive reward of $e_r = 1$ is awarded when the episode is successfully completed.
            Thus, the overall reward function for grasping and retraction is
            \begin{align}
                  r_{g} &= d_g + c_g + p_c + p_w + e_g \label{eq:rg},\\
                  r_{r} &= d_e + e_r \label{eq:rr}.
            \end{align}
            
        \subsubsection{Proximal Policy Optimization}
            \label{sec:methods:rl:ppo}
            Stable Baselines 3~\cite{raffinStableBaselines3ReliableReinforcement2021} and its implementation of \gls{ppo}~\cite{schulmanProximalPolicyOptimization2017} are utilized to train the agent.
            The agent is trained with a discount of $\gamma = 0.995$ and $\lambda_{GAE} = 0.95$ for a total of $10^7$ environment steps over $8$ parallel environments.
            The policy is updated after every $8 \times 128$ environment steps with a minibatch size of $256$ and $4$ epochs per \gls{ppo} iteration.
            Learning rate and clip ratio follow a linearly decreasing schedule starting at $0.1$ and $2.5\cdot10^{-4}$, respectively.
            \gls{ppo}'s ratio clip is set to $0.2$, the value and entropy loss coefficients are $0.5$ and $0.001$, respectively, and gradients are clipped to a maximum norm of $0.5$.
    
        \subsubsection{Agent Architecture}
            \label{sec:methods:rl:agent}
            The agent is split into policy and value estimation networks that do not share learnable parameters but are similar in architecture.
            Both networks consist of three convolutional layers with square kernel sizes $8, 4, 3$ and strides of $4, 2, 1$, followed by a fully connected layer with $512$ neurons.
            The policy network has a head with $3$ neurons for predicting the mean of a Gaussian distribution corresponding to the task-space velocities, and the value network has a head with $1$ neuron.
            The policy network also contains $3$ learnable log standard deviation parameters that do not depend on the input.
            ReLU non-linearities are applied after all layers except the final layer.
    
        \subsubsection{Curriculum Learning}
            \label{sec:methods:rl:curriculum}
            Curriculum learning~\cite{narvekarCurriculumLearningReinforcement2020} is an approach to gradually increase the difficulty of a learning environment in order to simplify learning complex tasks.
            This work employs curriculum learning to tune the weight of the collision and workspace violation terms of the reward function during grasping $r_g$~(Eq.~\ref{eq:rg}).
            The curriculum of the terms $p_c$ and $p_w$ of $r_g$ are defined as
            \begin{align}
                  p_{c} &= - w_c\frac{||\boldsymbol{x}_g - \boldsymbol{x}_t||_2}{||\boldsymbol{x}_{max} - \boldsymbol{x}_{min}||_2} \text{, if in collision} \\
                  p_{w} &= - w_w \text{, if action violates workspace}
            \end{align}
            with factors $w_c$ and $w_w$ linearly increasing from $0.0$ to $10$ and $0.2$, respectively, over $10^7$ environment steps.

    \section{Experimental Evaluation}
        \label{sec:experiments}
        The experimental setup consists of tissue represented by a rectangular silicone sheet attached to a board at a fixed set of attachment points, following the approach in~\cite{tagliabue2022deliberation}.
        The target is represented by a red circular marker on the board.
        The \gls{psm} on the \gls{dvrk} is equipped with a ProGrasp instrument, as illustrated in Fig.~\ref{fig:task}.
        An Intel RealSense D435 camera is used to capture monocular RGB images constituting the visual observations.
        
        \subsection{Unpaired Image-To-Image Translation}
            \label{sec:experiments:ui2i}
            The \gls{ui2i} models CUT, DCL, and CycleGAN are each trained with $7$ random seeds.
            Each training run is executed for 48 hours on four NVIDIA A100-40 GPUs with a minibatch size of $4$.
            The unpaired image dataset is created by performing grasping and retracting with different starting, grasping, and end points $82$ times under $3$ uncontrolled lighting conditions, for a total of $246$ \gls{tr} executions.
            RGB image observations are captured at \SI{30}{\Hz}.
            Data is captured on the real system and in simulation with a resolution of $256\times256$.
            This dataset is split randomly into $90\%$ for training and $5\%$ each for validation and testing.
        \subsection{Sim-To-Real Evaluation Scenarios}
            \label{sec:experiments:scenarios}
            \input{plots/scenarios}
            The experimental goal is to evaluate how well a policy that is trained on translated images in simulation performs on the real robotic setup.
            This is done without manually accounting for physical inaccuracies, such as modelling the robot dynamics, or environment conditions, such as changes in lighting, posing a more realistic and challenging task.
            
            The proposed approach is evaluated in four different scenarios as illustrated in Fig.~\ref{fig:scenarios}.
            The \textbf{first scenario}, denoted $\pi_s(O_s)$, evaluates the physical domain gap between simulation and reality by conditioning on observations $o_s$ from the simulation under state transitions $T_r$ from the real system.
            Policy $\pi_s$ is trained in simulation without image translation and then executed on the real system in a digital twin approach, where the gripper position in simulation is set to match the gripper position on the real system.
            The subscript $s$ indicates that this policy is conditioned on observations $o_s$ from simulation.
            The policy receives observations from simulation but predicts actions that are applied to the real system instead of the simulation.
            The \textbf{second scenario}, denoted $\pi_g(G(O_s))$, follows the same approach, but additionally translates simulation images during training and execution.
            Policy $\pi_g$ is trained on translated images $\hat{o}_r = G(o_s)$ in simulation as indicated by the subscript $g$.
            The \textbf{third scenario}, denoted $\pi_g(O_r)$, is the target scenario of the sim-to-real transfer.
            The same policy $\pi_g$ receives real images $o_r$ during execution.
            The \textbf{fourth scenario}, denoted $\pi_g(G(O_r))$, evaluates whether the image translation model can mitigate the influence of changes in lighting conditions on the real system.
            Images $o_r$ from the real system are translated by $G$ before being passed to the policy, exploiting the learned identity mapping of real images as described in Section~\ref{sec:methods:da}.
            
            Each scenario is executed from $32$ different starting positions.
            The starting positions are determined by dividing the robot's planar workspace into a $4\times4$ grid to generate $16$ starting positions on the XY-plane.
            Each starting position is executed on two starting heights ($z = \SI{60}{\mm}$ and $z = \SI{70}{\mm}$) to sample different heights within the allowed workspace.
            
        \subsection{Evaluation Metrics}
            \label{sec:experiments:metrics}
            Policy performance is evaluated based on trajectory outcome and quality metrics.
            This work considers four different trajectory outcomes: (1) \textit{success} if the policy completes both grasping and retracting phases, exposing the visual target as shown in Fig.~\ref{fig:task}; (2) \textit{partial success} if the policy completes the grasping phase but does not expose the visual target; (3) \textit{tissue stress}, where execution is aborted prematurely if the gripper applies excessive stress on the tissue through collisions before grasping; and (4) \textit{time out} if the policy fails to reach the grasping point within the time limit of $1\,000$ steps.
            The termination criterion of outcome (3) is triggered if the gripper moves more than \SI{8}{mm} laterally in collision.
            The evaluated trajectory quality metrics are the number of steps in collision and the absolute path length.

\section{RESULTS}
\input{plots/learning_curve}
\label{sec:results}
    \subsection{Unpaired Image-To-Image Translation}
        \label{sec:results:ui2i}
        \input{plots/gan_images.tex}
        The unpaired image dataset contains $278\,735$ images from the real system collected over $2.58$ hours and $108\,994$ images generated in simulation.
        Figure~\ref{fig:ui2i} (a) illustrates that both CUT and DCL successfully learn the image translation task, while CycleGAN does not, producing inconsistent visual features and spurious features.
        On visual assessment, however, DCL produces more consistent results for images from early steps in the retraction phase and is thus used as the image translation model $G$ for \gls{rl}.
        Figure~\ref{fig:ui2i} (b) illustrates a case where DCL successfully translates the simulation image, but CUT fails to correctly change the appearance of the gripper for the closed grasp.
        The manual inspection of $600$ randomly chosen images showed that $60\%$ for CycleGAN, $30.5\%$ for CUT, and $6.5\%$ for DCL of the translated images showed inconsistent or spurious visual features.

    \subsection{Training in Simulation}
        \label{sec:results:sim}
        Learning \gls{tr} on translated images in simulation requires roughly $2$ million environment steps as illustrated in Fig.~\ref{fig:learning_curve}.
        The policy learns to grasp successfully after roughly $200\,000$ environment steps, much faster than learning the retraction part of the task.
        The impact of curriculum learning can be seen in the number of environment steps spent in collision and workspace violation.
        The task is learned quickly, yet with many safety violations.
        After learning the task and progressively increasing the importance of safe actions, the unsafe parts of the trajectories decrease while task success continues to improve.
        The learning can be roughly separated into three phases as indicated in Fig.~\ref{fig:learning_curve}: the agent learning to grasp the tissue, the agent learning to retract the tissue, and the agent optimizing its learned behaviour to reduce episode length, collisions, and workspace violations.
        Both policies $\pi_s$ and $\pi_g$ achieve a task success rate of $100\%$ in simulation by the end of training.

    \subsection{Sim-To-Real Evaluation}
        \label{sec:results:sim2real}
        \input{plots/task_success}
        \input{plots/metrics_table}
        The trajectory outcomes and quality metrics as described in Section~\ref{sec:experiments:metrics} are presented in Fig.~\ref{fig:tasksuccess} and Tab.~\ref{tab:metrics}.
        Scenario $\pi_s(O_s)$, where the baseline policy $\pi_s$ is executed in simulation with the digital twin in the loop, was successful from all starting positions.
        Obtained results for scenario $\pi_g(G(O_s))$, where policy $\pi_g$ is executed on translated images with the digital twin in the loop, include three trajectories that were aborted due to excessive collisions, and four timed out without solving the grasping task.
        When the same policy $\pi_g$ receives real images for scenario $\pi_g(O_r)$, the observed success rate was further reduced to $16/32$.
        In contrast to the two previous experiments, three trajectories solved the grasping task, but were not successful in exposing the visual target.
        The final scenario $\pi_g(G(O_r))$, with policy $\pi_g$ executed on real images passed through the image translation model $G$, reaches a success rate of $13/32$.
        The amount of steps in collision increased over all four scenarios leading to an increase in the number of trajectories terminated due to excessive tissue stress.
        
        \input{plots/starting_positions}
        Figure~\ref{fig:startingpositions} illustrates the relation between starting positions and trajectory outcome for scenario $\pi_g(O_r)$ in more detail.
        Starting positions near the rear left corner $(x,y) = (-50.0,-86.0)$~mm  of the tissue tend to time out without grasping, while starting positions near the rear right corner $(x,y) = (50.0,-86.0)$~mm tend to terminate due to excessive collisions.

\section{DISCUSSION}
    \label{sec:discussion}
    
    The results of scenario $\pi_g(O_r)$ show that the policy trained in simulation on the transformed observation function $G(O_s)$ is able to perform the \gls{tr} task directly from raw camera images from the real observation function $O_r$.
    This indicates that the trained model $G$ is able to bridge the visual domain gap.
    The intuition for scenario $\pi_g(G(O_r))$ was that $G$ may mitigate the influence of changes in illumination by translating real images from different lighting conditions into a consistent appearance.
    Surprisingly, the policy performs worse in this scenario than the policy on real images directly.
    This also supports the claim that the image-to-image translation results in the loss of some image information necessary to compensate for the novel dynamics on the real robot.
    
    On the real evaluation setup, starting positions near the rear right corner collide excessively with the tissue.
    These starting positions were furthest away from the grasping point and thus also lead to the longest trajectories.
    Longer trajectories were observed to have more collisions, since the simplified collision model in simulation is somewhat more permissive than reality.
    Future work will investigate additional reward terms that encourage safe policies that are robust to changes in environment dynamics.
    The trajectory outcomes of evaluation scenario $\pi_s(O_s)$ show that policy $\pi_s$ is robust to the physical domain gap between simulation and the real system even though the simulation neither models measurement inaccuracies nor the dynamic behavior of the cable driven mechanism of the \gls{psm}.
    Both $\pi_s(O_s)$ and $\pi_g(G(O_s))$ are evaluated on observations from the same distribution as they were trained on, but under different dynamics through the digital twin.
    The drop in performance when comparing $\pi_s(O_s)$ to $\pi_g(G(O_s))$ shows that policy $\pi_g$ learned a behavior that does not translate well to the additional safety constraints, \ie trajectory terminated on excessive tissue stress, as described in Section~\ref{sec:experiments:metrics}.
    This result indicates that learning on translated images results in policies less robust to changes in dynamics.
    This may be due to the relative visual complexity of the translated images compared to the ones from simulation, which may make it more difficult for the policy to infer the environment's state from the observation.
    Preliminary tests with policies trained on a single \gls{gan} showed much lower success rates than the proposed method that used an ensemble of \glspl{gan}.
    This strengthens the idea that using an ensemble of \glspl{gan} may be interpreted as a form of domain randomization.
    Future work may further investigate how employing an ensemble of \glspl{gan} compares to classical visual domain randomization.
    
    The \gls{gan} used in this work was trained with a total of $246$ \gls{tr} trajectories and occasional random motions in the environment.
    This is substantially less data than state-of-the-art methods that achieve comparable task success in the real world, but require $10\,000$ task executions~\cite{hoRetinaganObjectawareApproach2021} and additional task specific auxiliary tasks or real world labels.
    Data collection for \gls{tr} task execution was straightforward since the required motion could be planned from predefined start and end points of the motion.
    Tasks where motion planning has to consider the elastic behavior and dynamics of the deformable object may complicate data collection.
    Simulation and evaluation setup were calibrated to have the same camera perspective and object positions.
    This choice was made to limit the image-to-image translation task to changes in image appearance~\cite{raoRlcycleganReinforcementLearning2020, hoRetinaganObjectawareApproach2021}.

    The imbalance of required training steps for learning the two-phased task may indicate that the initially learned features relevant for solving the grasp phase contradict the features relevant for the retraction phase, and that relearning features that can be generalized to both phases requires prolonged learning time.
    Additional challenges were encountered over the course of this work that are not explicitly described in this contribution, but should be mentioned to aid future work in the field.
    (1) Learning success in simulation was highly dependent on the camera perspective.
    Multiple different camera positions were evaluated and some, that often did not substantially differ to the human eye, were so detrimental to training, that the task was not learnable by the agent.
    (2) Reward normalization was essential for fast and repeatable training success.
    (3) Normalizing the real observations with running mean and standard deviation to mitigate the effects of changes in $O_r$ caused by changing lighting conditions is unlikely to yield better results as changes in lighting change more than just the brightness of an image.
    Different lighting conditions may also change the overall hue of an image as well as the presence of shadows.

    The achieved task success rate of $50\%$ is aligned with results obtained in state-of-the-art works for image-based sim-to-real transfer in robotics~\cite{openaiLearningDexterousInHand2019}, although we expect that task success could be further improved by including more sophisticated methods such as hierarchical \gls{rl} to learn a policy per phase, pretraining the agent with behavioral cloning, or including additional \gls{dr} as proposed in~\cite{jamesSimtorealSimtosimDataefficient2019}.
    We do not include these methods here to focus on \gls{da} rather than the absolute task success.
    Preliminary experiments with the same approach and hyperparameters on a different surgical robotic task, a Tissue Manipulation setup similar to~\cite{shin2019autonomous}, yield comparable results and a $63\%$ task success rate.
    In contrast to \gls{tr}, Tissue Manipulation is a robotic control task in which a visual marking on a deformable object must be aligned with an overlayed target location on the image observation.
    The manipulation is indirect because the deformable object is manipulated at a grasping point that does not coincide with the visual marking of interest.
    Thus, the policy must learn to model the deformation in order to align the marking with the target location.
    In comparison to~\cite{shin2019autonomous}, our method does not rely on an image-processing pipeline to generate state-observations and does not require training the policy on the real robotic system.
    We plan to extend these preliminary experiments to show that the approach can be transferred to other image-based tasks without changes to the pipeline.
    Moreover, initial tests were performed with \gls{dr} for the \gls{tr} task but did not yield noticeable benefits over naive sim-to-real transfer without \gls{da} and was thus not included as a separate experiment.

    Although the proposed pipeline is able to perform image-based sim-to-real transfer of a soft tissue manipulation task, some limitations remain that will be addressed in future work.
    In the context of surgical task learning, the method may be transferred to other tasks without modification and creation of auxiliary tasks for stabilization.
    The learned weights of the image translation model, however, are unlikely to perform successfully on new tasks.
    Despite requiring significantly less data compared to related works~\cite{hoRetinaganObjectawareApproach2021, raoRlcycleganReinforcementLearning2020}, it is necessary to record unpaired image data to train the translation models on a new task.
    From an image-to-image translation perspective however, the greatest limiting factor for practical application is that changes in camera perspective introduce additional changes in image content, not only in appearance.
    The presented approach thus relies on a precise calibration of camera positions between real system and simulation.
    Achieving view independence would increase the practical applicability of the approach for more realistic surgical scenarios where camera perspectives change over the course of the task and calibration between real setup and simulation is challenging.
    
    Building on the achieved results, future work will investigate the approach on setups with increased realism that include complex visual features, a broader physical domain gap, and non-calibrated camera perspectives.
    
\section{CONCLUSION}
    \label{sec:conclusion}
    This work is the first successful demonstration of sim-to-real transfer of a visuomotor policy in the context of robotic surgery.
    A data efficient approach for sim-to-real transfer through \gls{da} utilizing a contrastive \gls{gan} is presented.
    The policy is trained in a soft-body simulation on image observations and then transferred to a real robotic system without retraining the policy.
    Furthermore, the contrastive \gls{gan} approach does not require task specific auxiliary tasks, and requires much less data to train compared to most examples from literature.
    Evaluation of the policy on the real system demonstrates successful sim-to-real transfer for a \gls{tr} task and points out several critical challenges that must be addressed in future work.
    The successful transfer of an image based policy from simulation to a real robotic system for deformable object manipulation paves the way towards making \gls{rl} viable for robotic surgery and is a sizeable leap towards cognitive surgical robots.

\bibliographystyle{IEEEtran.bst}
\bibliography{references}

\end{document}

%% file: plots/overview.tex
\begin{figure}[tb]
    \parbox{\columnwidth}{
        \vspace{1.5mm}
        \centering
        \begin{tikzpicture}
        
        \draw [line width=1pt, dashed] (0, -2) -- (0, 3.2);
        \node (training) [anchor=center] at (-2, 3) {\Large Training};
        \node (evaluation) [anchor=center] at (2, 3) {\Large Evaluation};
        
        \node (policy training) at (-3, -2) {\Large $\pi$};
        \node (policy evaluation) at (3, -2) {\Large $\pi$};
        
        \node (sim image) at (-1, 0) {\includegraphics[width=1.3cm]{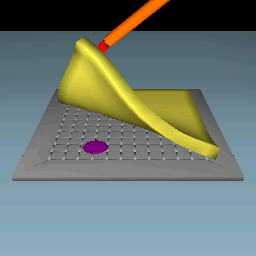}};
        \node (real image) at (1, 2) {\includegraphics[width=1.3cm]{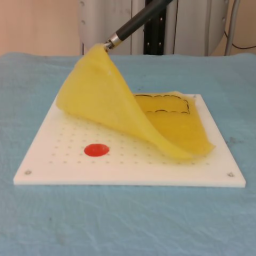}};
        \node (trans image) at (-1, 2) {\includegraphics[width=1.3cm]{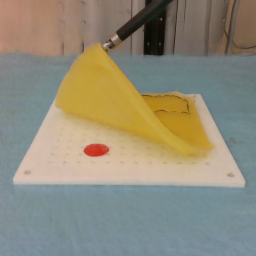}};
        
        \node (delta x train) at (-1, -1.5) {$\Delta(x, y, z)$};
        \node (delta x eval) at (1, -1.5) {$\Delta(x, y, z)$};
        
        \node (sofa) [draw, line width=1pt] at (-3, -1) {\includegraphics[width=1cm]{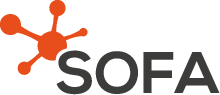}};
        \node (simulation) at (-3, -0.43) {\large Simulation};

        \draw [line width=1pt] (-3.5, 0.5) rectangle (-3.4, 1.5);
        \draw [line width=1pt] (-3.3, 0.7) rectangle (-3.2, 1.3);
        \node (G) at (-3.05, 1.5) {\large G};
        \draw [line width=1pt] (-3.1, 0.9) rectangle (-3.0, 1.1) node[fitting node] (gan) {};
        \draw [line width=1pt] (-2.9, 0.7) rectangle (-2.8, 1.3);
        \draw [line width=1pt] (-2.7, 0.5) rectangle (-2.6, 1.5);
        
        \node (real system) at (3.0, 0) {\includegraphics[width=1.5cm]{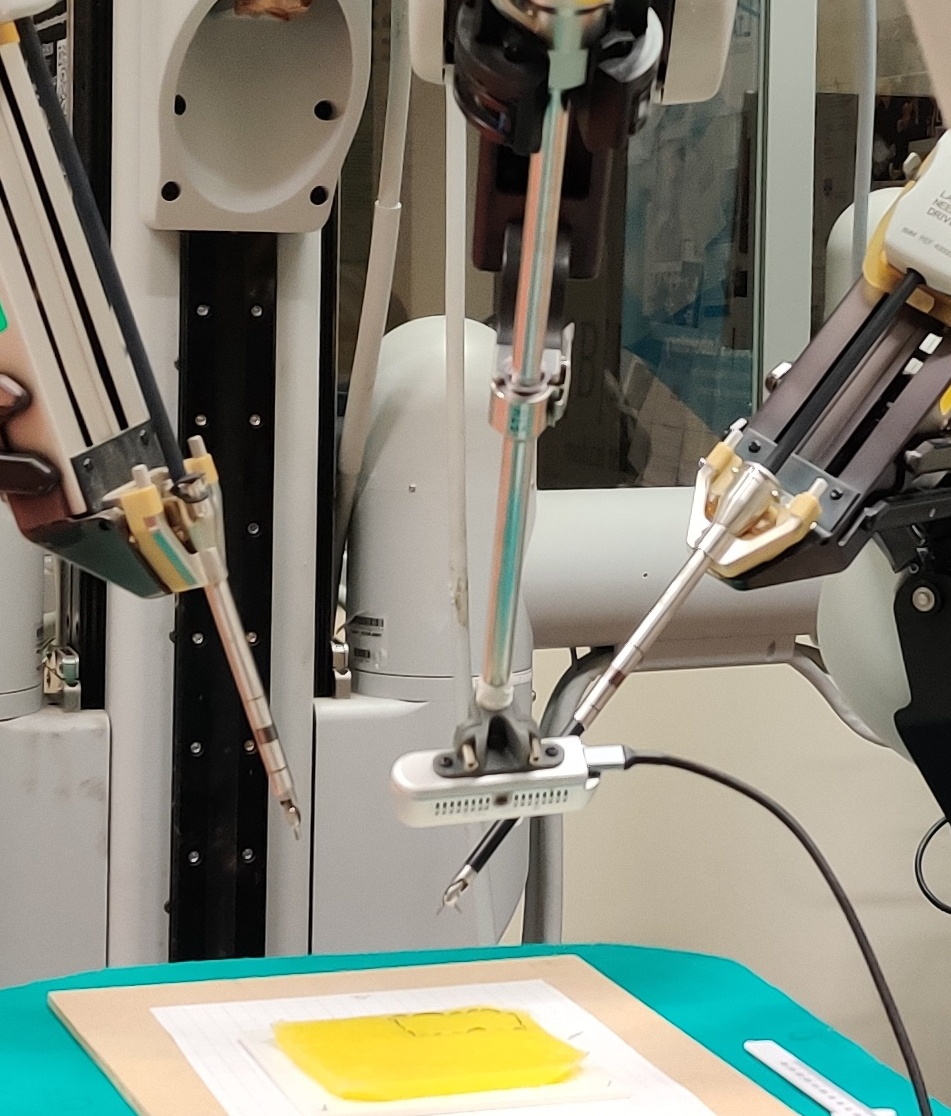}};
        \node (robot name) at (3.0, 1.15) {\large dVRK};

        \draw [->, line width=1pt] (G.north) -- (\the\tikz@lastxsaved, 2.6) -- ++(-1.2, 0) -- (\the\tikz@lastxsaved, -2) -- (policy training.west);
        \draw [->, line width=1pt] (policy training.north) -- (sofa.south);
        \draw [->, line width=1pt] (simulation.north) -- ++(0, 0.7);
        
        \draw [line width=1pt, dotted] (sim image.west) -- node[above] {$o_s$} (-3, \the\tikz@lastysaved);
        \draw [line width=1pt, dotted] (trans image.west) -- node[above] {$\hat{o}_r$} (-3, \the\tikz@lastysaved);
        \draw [line width=1pt, dotted] (delta x train.west) -- node[below] {$a$} (-3, \the\tikz@lastysaved);
        
        \draw [->, line width=1pt] (policy evaluation.north) -- (real system.south);
        \draw [->, line width=1pt] (robot name.north) -- (\the\tikz@lastxsaved, 2.6) -- ++(1.2, 0) -- (\the\tikz@lastxsaved, -2) -- (policy evaluation.east);
        
        \draw [line width=1pt, dotted] (real image.east) -- node[above] {$o_r$} (3, \the\tikz@lastysaved);
        \draw [line width=1pt, dotted] (delta x eval.east) -- node[below] {$a$} (3, \the\tikz@lastysaved);
        \end{tikzpicture}
    }
    \caption{Overview of training and evaluation settings. A policy $\pi$ is trained in simulation on translated observations $\hat{o}_r = G(o_s)$. During evaluation on the robotic system, the policy receives real image observations $o_r$. The actions $a$ of the policy are deltas in the gripper's Cartesian coordinates to solve a tissue retraction task.}
    \label{fig:training}
\end{figure}

%% file: plots/scenarios.tex
\newdimen\policyGEastX
\newdimen\policyGEastY
\newdimen\simulationWestX

\tikzset{every edge quotes/.style =
          { fill = white,
            execute at begin node = $,
            execute at end node   = $  }}
            
\begin{figure}[tb]
    \parbox{\columnwidth}{
        \vspace{1.6mm}
        \centering
        \begin{tikzpicture}
        
        \node [draw, line width=1pt, minimum width=1.5cm] (policy_s) at (-1, 1.5) {\large $\pi_s$};
        \node [draw, line width=1pt, minimum width=1.5cm] (policy_g) at (4, 1.5) {\large $\pi_g$};
        
        \node [draw, line width=1pt] (reality) at (4, 0) {\large Reality};
        \node [draw, line width=1pt] (simulation) at (-1, 0) {\large Simulation};

        \node [draw, line width=1pt] (gan) at (4, -1.5) {\large G};
        
        \pgfextractx{\policyGEastX}{\pgfpointanchor{policy_g}{east}}
        \pgfextracty{\policyGEastY}{\pgfpointanchor{policy_g}{east}}
        \pgfextractx{\simulationWestX}{\pgfpointanchor{simulation}{west}}
        
        \draw [PS, ->, line width=1pt, dashed] (simulation.north) to ["o_s"] (policy_s.south);
        \draw [TS, ->, line width=1pt, dash dot] (simulation.south east) to ["o_s"] (gan.north west);
        \draw [FS, ->, line width=1pt] (reality.south) to ["o_r"] (gan.north);
        \draw [TO, ->, line width=1pt, dotted] ([xshift=10pt]reality.south) -- ++(0.0, -0.5) to ["o_r"] ++(1.0, 0.0) -- ++(0.0, 2.15) -- (\policyGEastX, \the\tikz@lastysaved);

        \draw [PS, ->, line width=1pt, dashed] (policy_s.south east) to ["a"] (reality.north west);
        \draw [TS, ->, line width=1pt, dash dot] ([xshift=-10pt]policy_g.south) to ["a"] ([xshift=-10pt]reality.north);
        \draw [FS, ->, line width=1pt] (policy_g.south) to ["a"] (reality.north);
        \draw [TO, ->, line width=1pt, dotted] ([xshift=10pt]policy_g.south) to ["a"] ([xshift=10pt]reality.north);
        
        \path ([yshift=6pt, xshift=5pt]gan.east) -- node[FS] (text) {$G(o_r)$} ++(1.0, 0.0);
        \draw [FS, ->, line width=1pt] ([yshift=6pt]gan.east) -- (text.west) (text.east) -- ++(0.1, 0.0) -- (\the\tikz@lastxsaved, \policyGEastY) -- ([yshift=0pt]policy_g.east);
        \path ([yshift=-6pt, xshift=5pt]gan.east) -- node[TS] (text2) {$G(o_s)$} ++(1.0, 0.0);
        \draw [TS, ->, line width=1pt, dash dot] ([yshift=-6pt]gan.east) -- (text2.west) (text2.east) -- ++(0.3, 0.0) -- ++(0.0, 3.4) -- (\policyGEastX, \the\tikz@lastysaved);

        \draw [PS, ->, line width=1pt, dashed] ([yshift=4pt]reality.west) to ["z"] ([yshift=4pt]simulation.east);
        \draw [TS, ->, line width=1pt, dash dot] ([yshift=-4pt]reality.west) to ["z"] ([yshift=-4pt]simulation.east);
        
        \node (scenario 1) [anchor=west, xshift=10pt] at (-2.2, -1.0) {$\pi_s(O_s)$};
        \draw [PS, line width=1.5pt, dashed] (scenario 1.west)+(-0.3,0.0) to (scenario 1.west)+(-0.1,0.0);
        
        \node (scenario 2) [anchor=west, xshift=10pt] at (-0.5, -1.0) {$\pi_g(G(O_s))$};
        \draw [TS, line width=1.5pt, dash dot] (scenario 2.west)+(-0.3,0.0) to (scenario 2.west)+(-0.1,0.0);
        
        \node (scenario 3) [anchor=west, xshift=10pt] at (-2.2, -1.5) {$\pi_g(O_r)$};
        \draw [TO, line width=1.5pt, dotted] (scenario 3.west)+(-0.3,0.0) to (scenario 3.west)+(-0.1,0.0);
        
        \node (scenario 4) [anchor=west, xshift=10pt] at (-0.5, -1.5) {$\pi_g(G(O_r))$};
        \draw [FS, line width=1.5pt] (scenario 4.west)+(-0.3,0.0) to (scenario 4.west)+(-0.1,0.0);
        
        \end{tikzpicture}
    }
    \caption{Illustrated data flow for the four sim-to-real evaluation scenarios. All policy actions $a$ are executed on the real robot. For scenarios $\pi_s(O_s)$ and $\pi_g(G(O_s))$ the simulation is updated with the real robot states $s$ to generate observations $o_s$ from simulation. Scenarios $\pi_g(O_r)$ and $\pi_g(G(O_r))$ receive observations $o_r$ from the real system.}
    \label{fig:scenarios}
\end{figure}
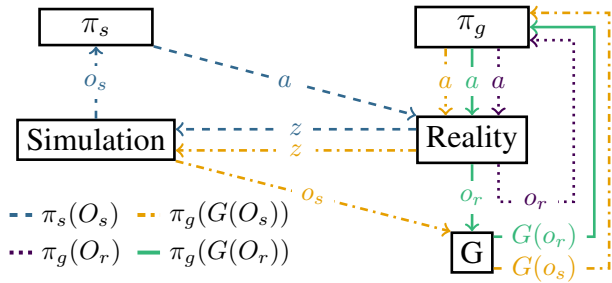

%% file: plots/learning_curve.tex
\begin{figure}[tb]
       \parbox{\columnwidth}{
         \vspace{1.6mm}
         \begin{tikzpicture}
                \begin{axis}[
                        axis x line=bottom,
                        axis y line=middle,
                        axis line style={-},
                        x label style={at={(axis description cs:0.5,-0.1)},anchor=north},
                        y label style={at={(axis description cs:-0.1,0.5)},rotate=90,anchor=south},
                        xlabel={\small Total Environment Steps},
                        ylabel={\small Return},
                        grid = major,
                        grid style={dashed, black!30},
                        xmax=4300000,
                        width=0.95\columnwidth,
                        height=0.5\columnwidth,
                        yticklabel style = {font=\scriptsize,xshift=0.5ex},
                        xticklabel style = {font=\scriptsize,yshift=0.5ex},
                        every x tick scale label/.style={at={(1,-0.1)},xshift=1pt,anchor=north,inner sep=0pt}
                     ]
                    \draw (axis description cs:0,1) -- (axis description cs:1,1);
                    \addplot[smooth, no markers, black, thick, tension=0.1] table[x=globalStep, y=return, col sep=comma] {plots/learningData.csv};
                    \label{return_plot}
                    \begin{scope}[on background layer]
                        \fill[TO,opacity=0.3] ({rel axis cs:0,0}) rectangle ({rel axis cs:0.04,1});
                        \fill[PS,opacity=0.3] ({rel axis cs:0.04,0}) rectangle ({rel axis cs:0.23,1});
                        \fill[FS,opacity=0.3] ({rel axis cs:0.23,0}) rectangle ({rel axis cs:1,1});
                    \end{scope}
                \end{axis}

                \begin{axis}[
                        axis x line=bottom,
                        axis y line=right,
                        axis line style={-},
                        xmax=4300000,
                        x label style={opacity=0},
                        y label style={at={(axis description cs:1.14,0.5)},rotate=0,anchor=south},
                        xlabel={},
                        ylabel={\small Steps},
                        enlarge y limits=true,
                        width=0.95\columnwidth, 
                        height=0.5\columnwidth,
                        legend style={fill opacity=0.8, draw opacity=1, text opacity=1, at={(1,0.6)}, anchor=east, nodes={scale=0.8, transform shape}, draw=none},
                        yticklabel style = {font=\scriptsize,xshift=-0.5ex},
                        xticklabel style = {opacity=0},
                     ]
                    \addplot[smooth, no markers, purple, thick, dash dot] table[x=globalStep,y=length,col sep=comma] {plots/learningData.csv};
                    \addlegendentry{\small{episode length}}
                    \addplot[smooth, no markers, OliveGreen, thick, dashed] table[x=globalStep,y=collision,col sep=comma] {plots/learningData.csv};
                    \addlegendentry{\small{steps in collision}}
                    \addplot[smooth, no markers, blue, thick, dotted] table[x=globalStep,y=workspace,col sep=comma] {plots/learningData.csv};
                    \addlegendentry{\small{steps in workspace violation}}
                    \addlegendimage{/pgfplots/refstyle=return_plot}\addlegendentry{\small{episode return}}
                \end{axis}
                
            \end{tikzpicture}
       }
       \caption{Smoothed learning curves for a training run of $\pi_g$. Average episode return, episode length, and steps in collision and workspace violation are shown over total steps in the learning environment. Three phases are identifiable during learning: when the agent is predominantly learning to grasp (\textcolor{TO}{purple}), when it is learning to retract (\textcolor{PS}{blue}), and when it is mainly optimizing to reduce episode length and collisions (\textcolor{FS}{green}).}
       \label{fig:learning_curve}
\end{figure}
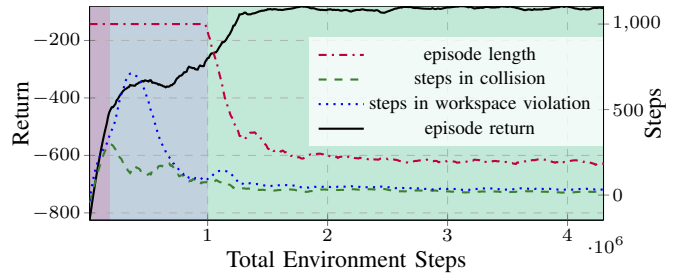

%% file: plots/gan_images.tex
\newdimen\floorY
\newdimen\ceilY
\newdimen\imageCeilY
\begin{figure*}
    \centering
    \vspace{1.5mm}
    \begin{tikzpicture}[align=center]
        \node[anchor=south west] (sim0) at (0, 6) {\includegraphics[height=2.0cm]{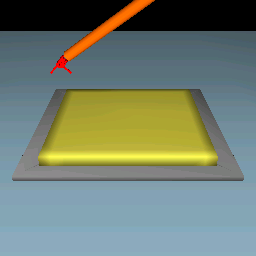}};
        \node[anchor=south west, below = 0 of sim0] (sim1) {\includegraphics[height=2.0cm]{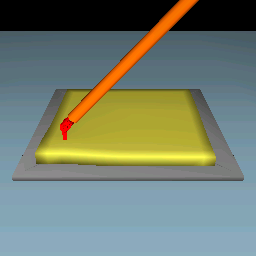}};
        \node[anchor=south west, below = 0 of sim1] (sim2) {\includegraphics[height=2.0cm]{images/gan/sim_0120.png}};
        \node[anchor=south, text depth=0pt, above = 0 of sim0] (text) {\large Simulation};
        \pgfextracty{\floorY}{\pgfpointanchor{sim2}{south}}
        \pgfextracty{\ceilY}{\pgfpointanchor{text}{north}}
        \pgfextracty{\imageCeilY}{\pgfpointanchor{sim0}{north}}
        
        \node[anchor=south west, right = 0 of sim0] (cg0) {\includegraphics[height=2.0cm]{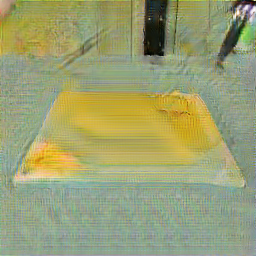}};
        \node[anchor=south west, below = 0 of cg0] (cg1) {\includegraphics[height=2.0cm]{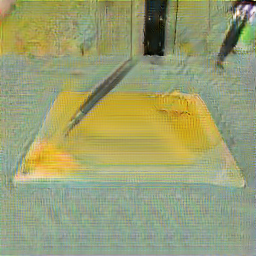}};
        \node[anchor=south west, below = 0 of cg1] {\includegraphics[height=2.0cm]{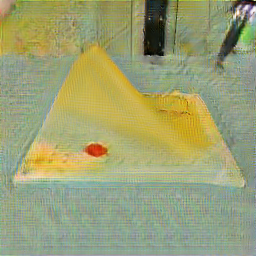}};
        \node[anchor=south, text depth=0pt, above = 0 of cg0] {\large CycleGAN};
        
        \node[anchor=south west, right = 0 of cg0] (cut0) {\includegraphics[height=2.0cm]{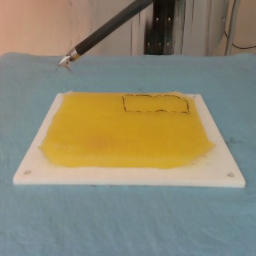}};
        \node[anchor=south west, below = 0 of cut0] (cut1) {\includegraphics[height=2.0cm]{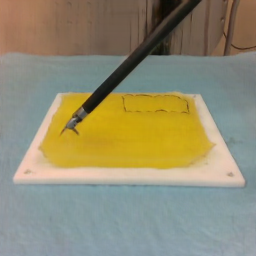}};
        \node[anchor=south west, below = 0 of cut1] (cut2) {\includegraphics[height=2.0cm]{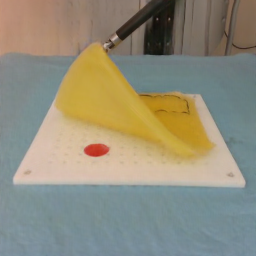}};
        \node[anchor=south text depth=0pt, above = 0 of cut0] {\large CUT};

        \node[anchor=south west, right = 0 of cut0] (dcl0) {\includegraphics[height=2.0cm]{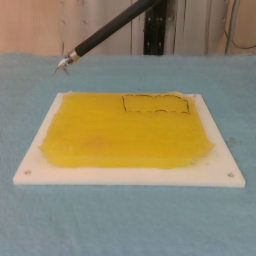}};
        \node[anchor=south west, below = 0 of dcl0] (dcl1) {\includegraphics[height=2.0cm]{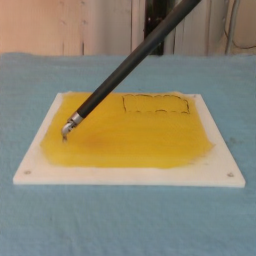}};
        \node[anchor=south west, below = 0 of dcl1] {\includegraphics[height=2.0cm]{images/gan/dcl_1649972301_0120.png}};
        \node[anchor=south text depth=0pt, above = 0 of dcl0] {\large DCL};

        \node[anchor=south west, right = 0 of dcl0] (real0) {\includegraphics[height=2.0cm]{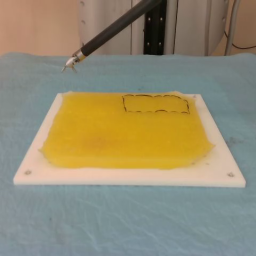}};
        \node[anchor=south west, below = 0 of real0] (real1) {\includegraphics[height=2.0cm]{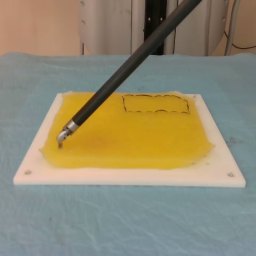}};
        \node[anchor=south west, below = 0 of real1] (real2) {\includegraphics[height=2.0cm]{images/gan/train_B_001190.png}};
        \node[anchor=south text depth=0pt, above = 0 of real0] {\large Real};
        
        \node[anchor=north west] (highlight cut) at (11.5, \imageCeilY) {\includegraphics[height=2.5cm]{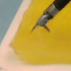}};
        \node[anchor=south west] (highlight dcl) at (11.5, \floorY) {\includegraphics[height=2.5cm]{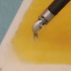}};
        \node[anchor=west] (highlight sim) at (14.3, 5) {\includegraphics[height=2.5cm]{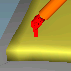}};
        
        \draw[->, line width=1pt] (highlight sim.north) -- node[above, shift={(0.5, 0.0)}] {\large CUT} (highlight cut.east);
        \draw[->, line width=1pt] (highlight sim.south) -- node[below, shift={(0.5, 0.0)}] {\large DCL} (highlight dcl.east);
        
        \draw[dashed, line width = 1pt] (11.35, \ceilY) -- (11.35, \floorY);
        \node[anchor=north, below = 0 of cut2] {(a)};
        \node[anchor=north, below = 0 of highlight dcl.south east] {(b)};

    \end{tikzpicture}
    \caption{Images from simulation and their translations by CycleGAN, CUT, and DCL, as well as real images (a) and a magnified view (b) on Simulation, CUT, and DCL images from the second row of (a).}
    \label{fig:ui2i}
\end{figure*}

%% file: plots/task_success.tex
\begin{figure}[tb]
    \parbox{\columnwidth}{
        \vspace{1.5mm}
        \begin{tikzpicture}
            \begin{axis}[
                ybar stacked,
                bar width=30pt,
                nodes near coords,
                enlarge x limits=0.15,
                width=1.05\columnwidth,
                height=0.6\columnwidth,
                ymin=0.0,
                ymax=33,
                legend style={at={(0.5,1.0)},anchor=south,legend columns=-1,draw=none,font=\small},
                ylabel={\small Number of Trajectories},
                y label style={at={(axis description cs:-0.0,0.5)}, anchor=south},
                symbolic x coords={$\pi_s(O_s)$, $\pi_g(G(O_s))$, $\pi_g(O_r)$, $\pi_g(G(O_r))$},
                xtick=data,
                y tick style={draw=none},
                yticklabels={,,},
                x tick style={draw=none},
                x tick label style={rotate=0,anchor=north, font=\small},
                every node near coord/.append style={font=\boldmath},
                ]
                \addplot+[ybar, Black, fill=FS, draw=Black] plot coordinates {($\pi_s(O_s)$, 32) ($\pi_g(G(O_s))$, 25) ($\pi_g(G(O_r))$, 13) ($\pi_g(O_r)$, 16)};
                
                \addplot+[ybar, Black, fill=PS, draw=Black, text=white] plot coordinates {($\pi_s(O_s)$, 0) ($\pi_g(G(O_s))$, 0) ($\pi_g(G(O_r))$, 0) ($\pi_g(O_r)$, 3)};
                
                \addplot+[ybar, Black, fill=TS, draw=Black] plot coordinates {($\pi_s(O_s)$, 0) ($\pi_g(G(O_s))$, 3) ($\pi_g(G(O_r))$, 15) ($\pi_g(O_r)$, 9)};
                
                \addplot+[ybar, Black, fill=TO, draw=Black, text=white] plot coordinates {($\pi_s(O_s)$, 0) ($\pi_g(G(O_s))$, 4) ($\pi_g(G(O_r))$, 4) ($\pi_g(O_r)$, 4)};
                
                \legend{\strut success, \strut partial success, \strut tissue stress, \strut time out}
            \end{axis}
        \end{tikzpicture}
    }
    \caption{Trajectory results for all four experiment cases split into their outcomes. Possible trajectory outcomes are task failure due to time out or excessive tissue stress, partial task success if grasping is successful but retraction failed, and success if the target was visible at trajectory end.}
    \label{fig:tasksuccess}
\end{figure}
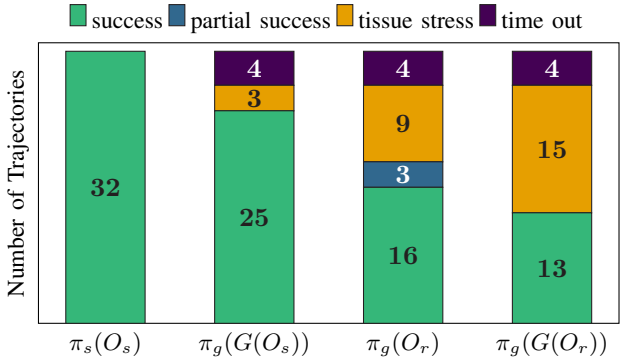

%% file: plots/metrics_table.tex
\begin{table}
    \centering
    \vspace{1.5mm}
    \caption{Evaluation results for the scenarios defined in Section~\ref{sec:experiments:scenarios} with regard to the metrics defined in Section~\ref{sec:experiments:metrics}.}
    \begin{tabular*}{\columnwidth}{ @{\extracolsep{\fill}} lccr}
        \toprule
        Scenario         & Success Rate   & Path Length   & Collisions \\
        \midrule
        $\pi_s(O_s)$     & $32/32$        & \SI{210}{\mm} & $2.93$ steps  \\
        $\pi_g(G(O_s))$  & $25/32$        & \SI{235}{\mm} & $12.16$ steps \\
        $\pi_g(O_r)$     & $16/32$        & \SI{219}{\mm} & $14.56$ steps \\
        $\pi_g(G(O_r))$  & $13/32$        & \SI{210}{\mm} & $18.08$ steps \\
        \bottomrule
    \end{tabular*}
    \label{tab:metrics}
\end{table}

%% file: plots/starting_positions.tex
\def\layerOneColors{{
"FS",
"FS",
"TO",
"TO",
"FS",
"FS",
"FS",
"TS",
"FS",
"FS",
"TS",
"FS",
"FS",
"TS",
"TS",
"TS"
}}

\def\layerTwoColors{{
"TO",
"FS",
"PS",
"TO",
"FS",
"FS",
"TS",
"PS",
"TS",
"FS",
"PS",
"FS",
"FS",
"FS",
"TS",
"TS"
}}

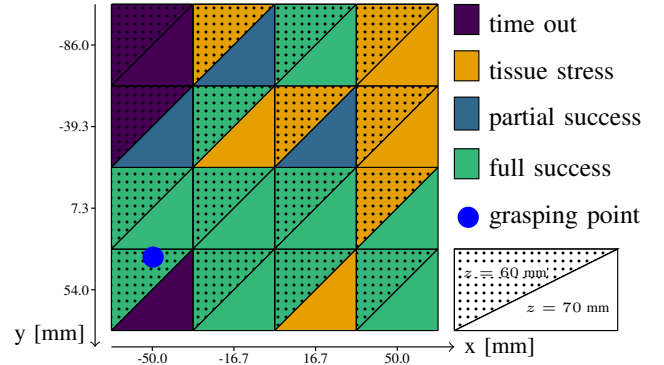
\begin{figure}[tb]
    \parbox{\columnwidth}{
        \vspace{1.5mm}
        \resizebox{\columnwidth}{!}{
        \begin{tikzpicture}[node distance=0.1]
            \foreach \x in {0,...,3}
                \foreach \y in {0,...,3} 
                    {
                        \pgfmathsetmacro{\colorIndex}{int(\x * 4 + \y )};
                        \pgfmathsetmacro{\currentColorLayerOne}{\layerOneColors[\colorIndex]};
                        \pgfmathsetmacro{\currentColorLayerTwo}{\layerTwoColors[\colorIndex]};
                        \pgfmathsetmacro{\layerOneIndex}{int(\colorIndex + 1)};
                        \pgfmathsetmacro{\layerTwoIndex}{int(\colorIndex + 17)};
                        \draw [fill=\currentColorLayerOne, rounded corners=0.1mm] (\x,\y) -- (\x+1,\y+1) -- (\x,\y+1) -- cycle;                        
                        \draw [pattern=dots, pattern color=black ,rounded corners=0.1mm] (\x,\y) -- (\x+1,\y+1) -- (\x,\y+1) -- cycle;
                        \draw [rounded corners=0.1mm, fill=\currentColorLayerTwo] (\x,\y) -- (\x+1,\y) -- (\x+1,\y+1) -- cycle;
                        \node (T\layerOneIndex) [anchor=center] at ({(3*\x + 1)/3}, {(3*\y + 2)/3}) {};
                        \node (T\layerTwoIndex) [anchor=center] at ({(3*\x + 2)/3}, {(3*\y + 1)/3}) {};
                    } 
            
            \draw[->] (-0.2, 4) -- (-0.2, -0.2);
            \node [anchor=south east] at (-0.2, -0.3) {\footnotesize y\ [mm]};
            \foreach \z [count=\zi from 0] in {-86.0, -39.3, 7.3, 54.0}
                {
                    \node [anchor=east] at (-0.15, 3.5 - \zi) {\tiny \z};
                    \draw (-0.25, 3.5 - \zi) -- (-0.2, 3.5 - \zi);
                }
            
            \draw[->] (0.0, -0.2) -- (4.2, -0.2);
            \node [anchor=west] at (4.2, -0.2) {\footnotesize x\ [mm]};
            \foreach \x [count=\xi from 0] in {-50.0, -16.7, 16.7, 50.0}
                {
                    \node [anchor=north] at (\xi + 0.5, -0.15) {\tiny \x};
                    \draw (\xi + 0.5, -0.2) -- (\xi + 0.5, -0.25);
                }

            \draw [pattern=dots, pattern color=black] (4.2,0) -- (6.2,1) -- (4.2,1) -- (4.2,0);
            \node [fill=white, inner sep=0.1,outer sep=0.1] (Y1) [anchor=center] at ({(2*4.2+6.2)/3-0.05}, {(2*1)/3+0.05}) {\tiny $z=60$ mm};
            \draw (4.2,0) -- (6.2,0) -- (6.2,1) -- (4.2,1) -- (4.2,0) -- (6.2,1);
        
            \node (Y2) [anchor=center] at ({(4.2+2*6.2)/3+0.05}, {(1)/3-0.05}) {\tiny $z=70$ mm};
            
            \node (legend) at (4.5, 4) {};
            
            \node (A) [below of=legend, anchor=north west] {\small time out};
            \draw [fill=TO] (A.south west)+(-0.3,0.1) rectangle (A.north west)+(-0.1,0.0);
            
            \node (B) [below = of A.south west, anchor=north west] {\small tissue stress};
            \draw [fill=TS] (B.south west)+(-0.3,0.1) rectangle (B.north west)+(-0.1,0.0);
            
            \node (C) [below = of B.south west, anchor=north west] {\small partial success};
            \draw [fill=PS] (C.south west)+(-0.3,0.13) rectangle (C.north west)+(-0.1,0.0);
            
            \node (D) [below = of C.south west, anchor=north west] {\small full success};
            \draw [fill=FS] (D.south west)+(-0.3,0.1) rectangle (D.north west)+(-0.1,0.0);
            
            \node (E) [below = of D.south west, anchor=north west] {\small grasping point};
            \draw [fill=blue, draw=none] (E.south west)+(-0.13,0.26) circle (0.13);

            \draw [fill=blue, draw=none](0.51, 0.9) circle (0.13);

        \end{tikzpicture}
        }
    }
    \caption{Trajectory outcome of case $\pi_g(O_r)$ for 16 evaluated starting positions on the XY-plane. Each starting position was executed on two starting heights ($z = \SI{60}{\mm}$ and $z = \SI{70}{\mm}$).}
    \label{fig:startingpositions}
\end{figure}